\documentclass[letterpaper, 10 pt, conference]{ieeeconf}  

\IEEEoverridecommandlockouts                              

\makeatletter
\let\NAT@parse\undefined
\makeatother
\usepackage{graphics} 
\usepackage{graphicx}
\usepackage{grffile} 
\usepackage{amsmath} 
\usepackage{amssymb}  
\usepackage{color}
\usepackage{mathtools} 
\usepackage{booktabs,ragged2e}
\usepackage{breqn}
\usepackage{textgreek}
\usepackage{arydshln}
\usepackage[flushleft]{threeparttable}
\usepackage{tabularx,booktabs}
\usepackage{graphics}
\usepackage{amsmath}
\usepackage{gensymb}
\usepackage[normalem]{ulem} 
\usepackage{pifont}
\usepackage{booktabs}
\usepackage{multirow}

\usepackage{lineno}
\usepackage{cite}
\usepackage{bm}
\usepackage{url}

\usepackage[noend]{algpseudocode}
\usepackage{algorithm}
\usepackage{makecell}
\usepackage[square, comma, numbers,sort&compress]{natbib}
\usepackage{multirow}
\let\labelindent\relax
\usepackage{enumitem}
\usepackage[table,dvipsnames]{xcolor}
\usepackage[pagebackref=true,breaklinks=true,colorlinks,bookmarks=false]{hyperref}
\usepackage{comment}
\usepackage{svg}
\hypersetup{
linkcolor=BrickRed
,citecolor=Green
,filecolor=Mulberry
,urlcolor=NavyBlue
,menucolor=BrickRed
,runcolor=Mulberry
,linkbordercolor=BrickRed
,citebordercolor=Green
,filebordercolor=Mulberry
,urlbordercolor=NavyBlue
,menubordercolor=BrickRed
,runbordercolor=Mulberry
}

\usepackage[disable]{todonotes}
\usepackage[normalem]{ulem} 

\title{\LARGE \bf
POE: Acoustic Soft Robotic Proprioception for Omnidirectional End-effectors
}

\author{\small Uksang Yoo \textsuperscript{1,\ding{41}}
, Ziven Lopez\textsuperscript{2} , Jeffrey Ichnowski\textsuperscript{1*}, and Jean Oh\textsuperscript{1*}
\thanks{\ding{41} Corresponding author.}%
\thanks{* Equal contribution}%
\thanks{$^{1}$Robotics Institute, Carnegie Mellon University, Pittsburgh, PA 15213, USA
	{\tt\small \{uyoo, jichnows, hyaejino\}@andrew.cmu.edu}}%
\thanks{$^{2}$Northeastern University, Boston, MA, 02115, USA
{\tt\small \{lopez.z\}@northeastern.edu}}
\thanks{The work is supported by NSF Graduate Research Fellowship under Grant No. DGE2140739.}
}

\begin{document}

\maketitle
\thispagestyle{empty}
\pagestyle{empty}
\renewcommand{\baselinestretch}{0.979} 

\begin{abstract}
Soft robotic shape estimation and proprioception are challenging because of soft robot's complex deformation behaviors and infinite degrees of freedom. A soft robot's continuously deforming body makes it difficult to integrate rigid sensors and to reliably estimate its shape. In this work, we present Proprioceptive Omnidirectional End-effector (POE), which has six embedded microphones across the tendon-driven soft robot's surface. We first introduce novel applications of previously proposed 3D reconstruction methods to acoustic signals from the microphones for soft robot shape proprioception. To improve the proprioception pipeline's training efficiency and model prediction consistency, we present POE-M. POE-M first predicts key point positions from the acoustic signal observations with the embedded microphone array. Then we utilize an energy-minimization method to reconstruct a physically admissible high-resolution mesh of POE given the estimated key points. We evaluate the mesh reconstruction module with simulated data and the full POE-M pipeline with real-world experiments. We demonstrate that POE-M's explicit guidance of the key points during the mesh reconstruction process provides robustness and stability to the pipeline with ablation studies. POE-M reduced the maximum Chamfer distance error by 23.10 $\%$ compared to the state-of-the-art end-to-end soft robot proprioception models and achieved 4.91 mm average Chamfer distance error during evaluation. 


\end{abstract}

\renewcommand{\baselinestretch}{0.979}

\section{Introduction}
Soft robots have advantages in applications such as fruit harvesting \cite{j_f_elfferich_soft_2022, a_l_gunderman_tendon-driven_2022}, food packaging \cite{low_sensorized_2022}, minimally invasive surgery \cite{runciman_soft_2019, k_-w_kwok_soft_2022}, and human-wearable robot interactions \cite{walsh_human---loop_2018,m_zhu_soft_2022}. Soft robots' constituent deformable and elastic material make them well-suited for handling delicate objects and preventing destructive interactions \cite{watanabe_survey_2017}. Furthermore, recent works have demonstrated that the compliance of the soft robots makes them more robust in contact-rich control tasks \cite{bhatt_surprisingly_2021, sieler_dexterous_2023}.

Soft robotic shape representation and estimation, also known as proprioception, are challenges rooted in soft robots' inherent deformability and underactuated high degrees of freedom \cite{wang_real-time_2019, yoo_toward_2023}. Without proprioception, soft robot's shape and state must be observed externally and limits the robot's workspace to the external sensor's field of perception and is subject to occlusion \cite{bruder_koopman-based_2021, haggerty_control_nodate}. Prior research has often relied on fitting low degree-of-freedom shape primitives, such as constant curvature curves to represent the centerline of the robot \cite{u_yoo_analytical_2021, r_szasz_modeling_2022}. Such approaches, however, filter out any complex deformation behaviors of the soft robots by reducing the independent degrees of freedom of the system.

\begin{figure}
    \centering
    \includegraphics[width=1.\linewidth]{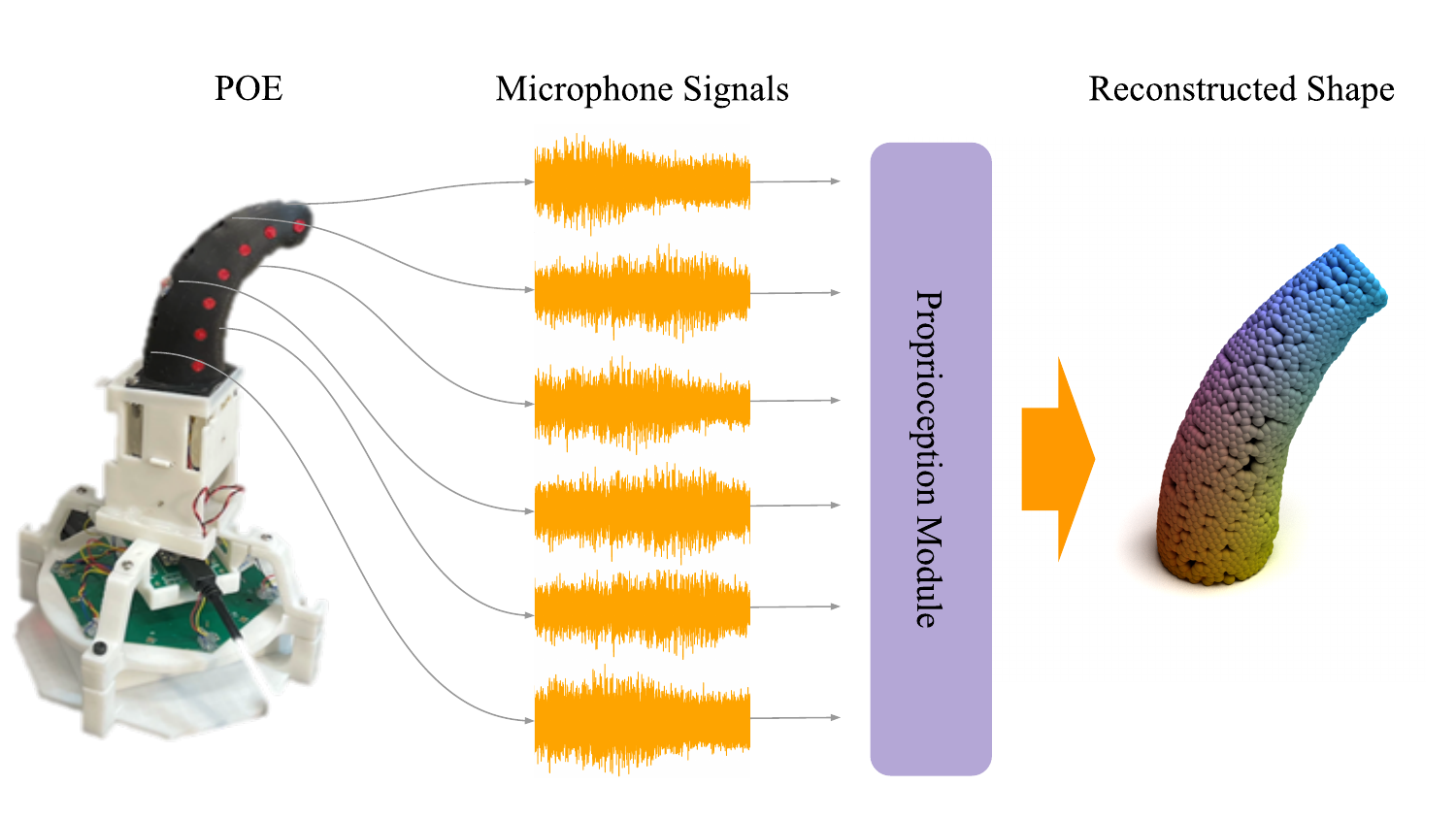}
    \caption{ Overview of the proposed pipeline for Proprioceptive Omnidirectional End-effector (POE) acoustic soft robotic proprioception. We obtain acoustic signals from our novel tendon-driven soft robot POE with six embedded microphones (left). We then feed the extracted acoustic features into our proprioception modules that are able to reconstruct a high degree of freedom shape of POE (right).}
    \label{fig:teaser}
    \vspace{-5mm}
\end{figure}

Using internally embedded cameras can help to capture the complex deformation behavior of the soft robots if the internal lighting and spatial conditions allow for vision-based perception~\cite{wang_real-time_2019,a_zhang_vision-based_2022}. In these applications, vision provided immensely rich observations of complex soft surface deformations \cite{wang_real-time_2019}. However, vision-based approaches require specially designed robots with a sufficiently large cavity to limit self-occlusion. Additionally, they generally suffer from difficulty in scaling, which limits their generalizability to different robot morphologies~\cite{yoo_toward_2023}. Furthermore, these methods generally utilize vision models that require a large training dataset of image-shape pairs. Previous works in vision-based soft robotic proprioception have also noted limitations to applicability. 

Acoustic sensing is an attractive alternative for soft robot proprioception applications because microphones can be small, scaled up to an array to cover a larger space, and be easily installed. To investigate acoustic proprioception for soft robots, we present acoustically Proprioceptive Omnidirectional End-effector (POE). POE has an array of embedded microphones to observe changes in the acoustic propagation properties of the soft robot with deformation. We demonstrate with baseline reconstruction pipelines based on K-nearest neighbors (KNN) and an encoder-decoder network inspired by DeepSoRoNet~\cite{wang_real-time_2019, yoo_toward_2023} that signals from embedded microphone arrays can be utilized to reconstruct the shape of the robot reliably in diverse loading conditions. Furthermore, we introduce POE-M, a pipeline that first predicts key points that guide the mesh reconstruction module toward full POE shape estimation. The real-world experiments with POE showed that POE-M can reconstruct the current state of the soft robot with 23.10$ \% $ lower maximum Chamfer distance error compared to learning to directly reconstruct the point cloud.

This paper makes the following contributions:
\begin{enumerate}
    \item Design and evaluation of a tendon-driven soft robot with embedded microphones and an active sound source,
    \item POE-M pipeline which approximates key points from acoustic signals and reconstructs the full watertight mesh of the deformed robot with energy-minimization method,
    \item Acoustic-shape dataset for POE with a thorough evaluation of the capabilities and limitations of the proposed pipelines.
\end{enumerate}

\renewcommand{\baselinestretch}{0.979} 
\section{Related Work}

\begin{figure}
    \centering
    \includegraphics[width=1.1\linewidth]{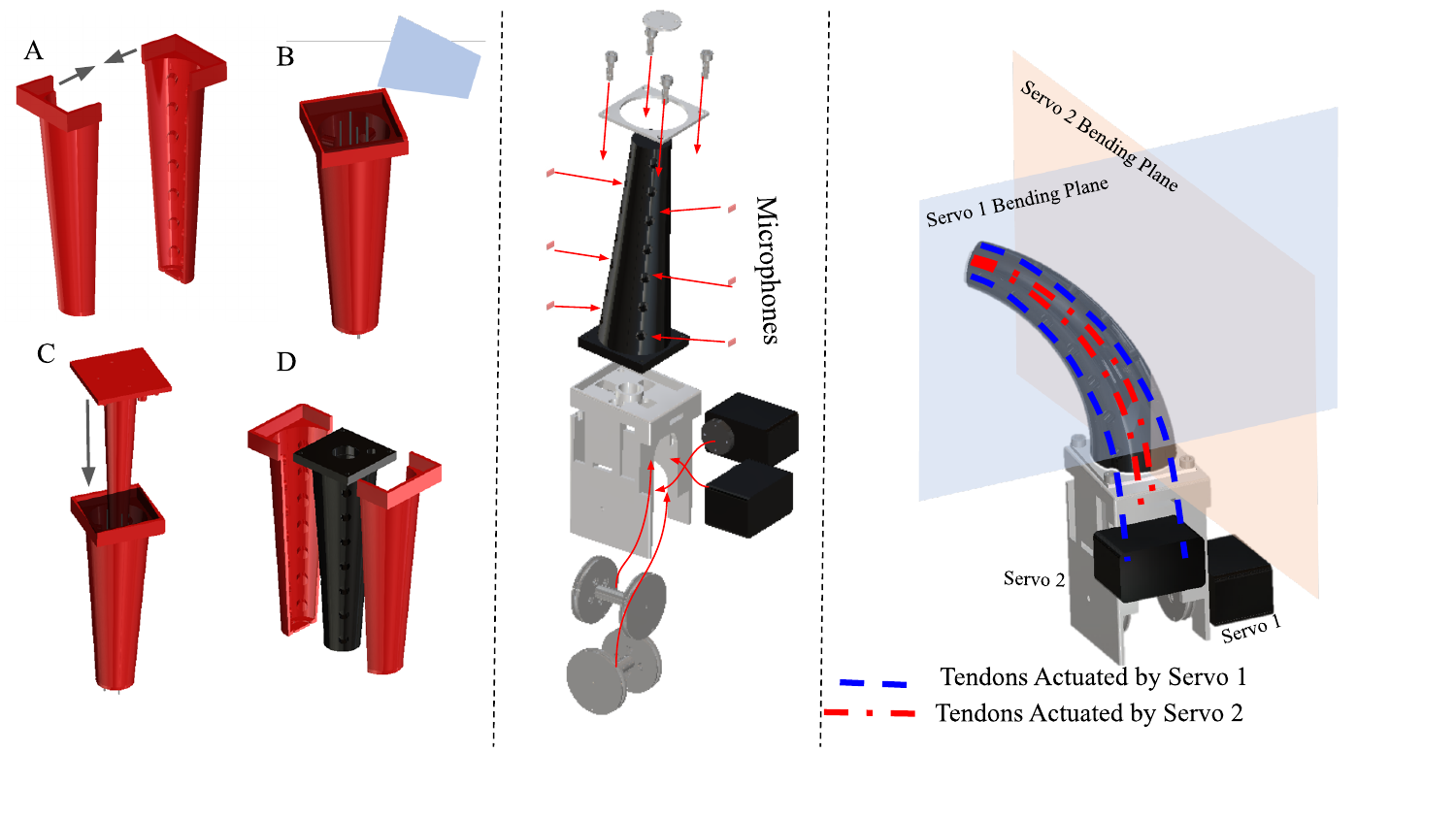}
    \caption{POE is a tendon-driven soft robot fabricated with a single molding step. \textbf{Left column:} The three-part mold is assembled with four metal rods (B) before the uncured silicone is poured in to create channels for tendons routing. The last part of the mold is then inserted to make the central conical cavity (C). After curing, the mold is disassembled to get the POE finger. \textbf{Middle column:} the finger secured into rest of POE assembly with two servo motors for tendon actuation. \textbf{Right column:} each servo controls POE movement in its perpendicular plane of bending, enabling POE to bend toward any direction. }
    \label{fig:design}
    \vspace{-5mm}
\end{figure}

\subsection{Soft Robotic Manipulators}
Design and development of soft-robotic manipulators are active fields of research with growing interest~\cite{rus_design_2015, hawkes_hard_2021}. The mode of actuation for soft robots varies widely from dielectric artificial muscle actuation to pneumatic. Pneumatically actuated and tendon-driven soft robots generally only require readily available materials that are easy to assemble \cite{u_yoo_analytical_2021, j_fras_instant_2020}. The relative fabrication simplicity of pneumatic and tendon-driven soft robots have made them popular methods of soft-robotic actuation~\cite{yasa_overview_2023}. 


\subsection{Soft Robotic Proprioception}

To address challenges in soft robot state estimation and proprioception, researchers have proposed various methods to directly measure deformations in a soft body~\cite{wang_toward_2018}. A popular approach is to embed flexible and elastic strain sensors along the length of the robot \cite{v_wall_method_2017,tapia_makesense_2020, so_shape_2021}. However, implementing these direct strain sensing methods can be difficult and expensive. Direct strain measurement methods generally require specialized sensors, hardware, and domain expertise to fabricate \cite{tapia_makesense_2020,zhao_optoelectronically_2016}. Furthermore, the soft robot design complexity grows substantially with an increasing number of these sensing elements, requiring robot-specific sensor placement optimization~\cite{v_wall_multi-task_2019}. These works have also focused on simplifying the shape representation to limited degree of freedom representations, often reducing the shape of the robot to a single curvature curve~\cite{v_wall_method_2017}.

Indirect methods of soft robotic proprioceptive sensing do not measure local strain and deformation directly. A popular approach within this family of methods is embedding a camera inside of the soft robot to observe deformations of the internal surface of the robot and infer the external shape of the robot~\cite{wang_real-time_2019,hofer_vision-based_2021, yoo_toward_2023,she_exoskeleton-covered_2020}. Because vision provides rich observations of the state of the robot, some previous works were compatible with a high degree-of-freedom soft-robot shape representations that capture the complex deformation behaviors better than low degree-of-freedom representations such as constant curvature models ~\cite{wang_real-time_2019}. These methods generally rely on an encoder-decoder neural network architecture where a latent code representation of the robot is learned which is then decoded to the full robot shape represented by point clouds. In the process, intuitive physics-based constraints are lost and must be learned implicitly from examples, requiring a large training dataset of image-shape pairs to consistently yield physically admissible soft-robot shape predictions~\cite{yoo_toward_2023}. The proposed POE-M approach addresses these challenges with stronger supervision by predicting key point movements on the mesh and framing the shape decoding task as a mesh energy optimization problem~\cite{sanchez-gonzalez_graph_2018}.
\begin{figure}
\vspace*{-15 pt}
    \centering
    \includegraphics[width=1.4\linewidth]{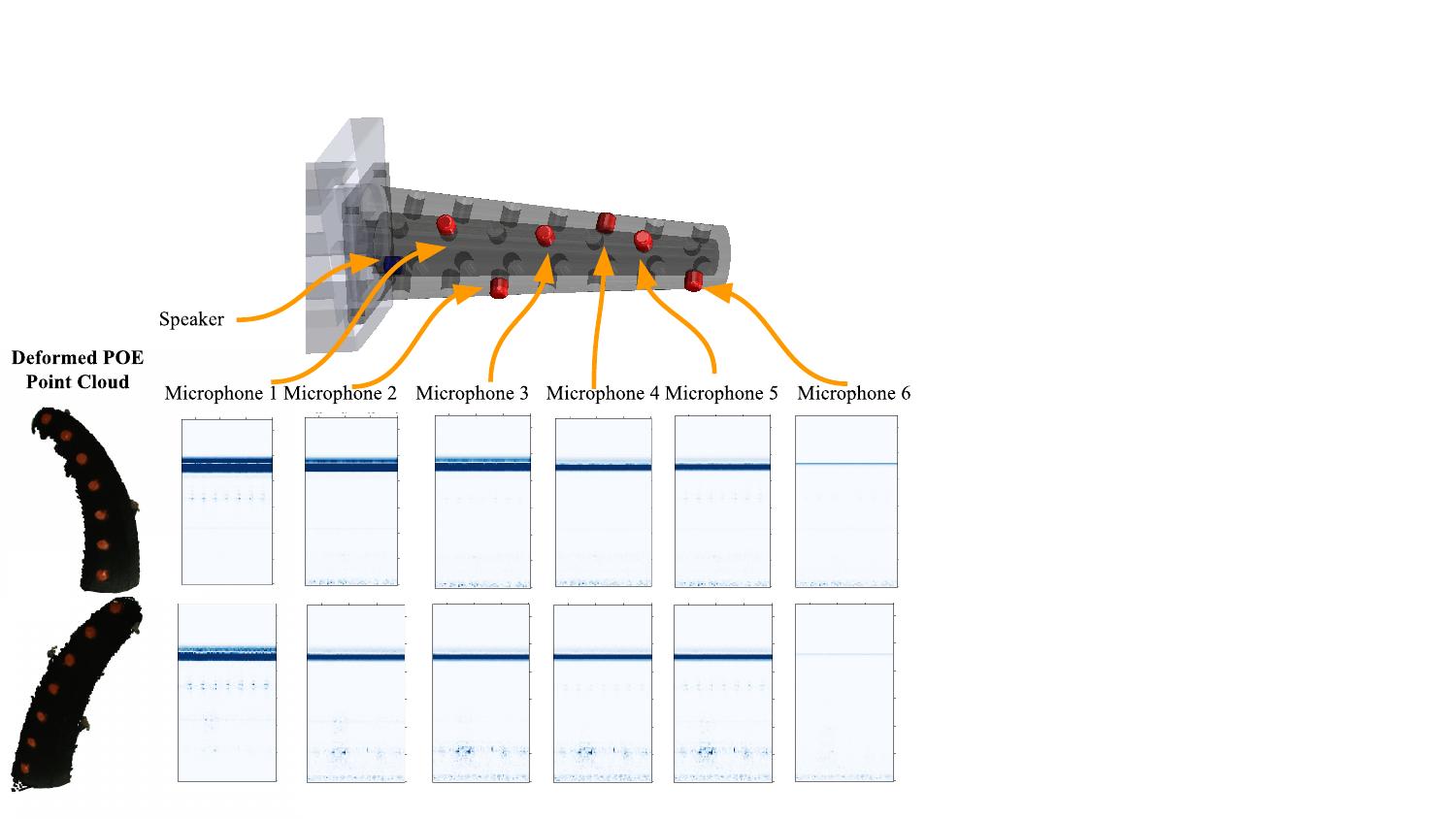}
    \caption{Spectogram visualizations from the six embedded microphones in two different shape configurations. The signal magnitudes change from one shape to another as we observe from the two rows of spectrograms. Time-varying features are ignored in the presented pipelines by averaging over the recording. }
    \label{fig:microphones}
\vspace*{-15pt}
\end{figure}
\subsection{Acoustic Sensing for Soft Robots}
Arrays of microphones can be embedded into soft bodies to provide indirect contact sensing in soft bodies~\cite{park_biomimetic_nodate}. For soft robots, acoustic sensing is particularly advantageous because low-cost easily-embedded miniature microphones are readily available~\cite{g_zoller_active_2020}. Furthermore, by using the soft robot's deformable body as the medium for propagation of acoustic vibrations from an active sound source, microphones can detect changes to the material state such as strain~\cite{wall_virtual_2022}. Previous works have demonstrated that a microphone embedded in the soft robots can be used for various applications ranging from contacting object material property classification, contact position estimation, temperature regression, and braille letter classification~\cite{g_zoller_active_2020,wall_virtual_2022,wall_passive_2023}. The previous approaches generally relied on a single embedded microphone and framed the tasks as a coarse classification or low degree-of-freedom regression problem~\cite{wall_virtual_2022}. To the best of our knowledge, this work presents the first multi-microphone high-fidelity shape reconstruction of the robot with acoustic signals.

\renewcommand{\baselinestretch}{0.979}

\section{Methods}
The following sections outline the methods implemented in this work toward developing POE and associated methods for acoustic proprioception.
\subsection{Design and Fabrication of POE}

We chose to develop POE with tendon-actuation for simplicity. POE is actuated by four tendons that are each anchored at the tip of the finger and tensions are applied with two servos (Dynamixel, XC330-M288-T). POE's pulley design allows each servo motor to control the robot pose in a bending plane as shown in Fig.~\ref{fig:design}. The combination of the two servos can actuate POE to move its fingertip to any point in its semi-hemisphere workspace without contact (Fig.~\ref{fig:design}).

POE's soft finger is molded from a 3-part mold with silicone rubber (Smooth-On, Inc. Dragon Skin\textsuperscript{TM} 20). The mold has four narrow cavities where the nylon tendons are inserted through after curing. POE has a central conical cavity that allows it to bend without buckling. The overall length of POE is 110.0 mm. Six miniature microphones (ReSpeaker, Circular Array) are inserted into the preset cavities and secured with silicone adhesive. The microphone wires are channeled through the conical cavity. The base of POE's molded finger has an embedded miniature speaker (Knowles, RAB) to act as an active sound source. The acoustic signals that POE microphone arrays observe when POE is deformed into different shapes are shown in Fig.~\ref{fig:microphones} with their corresponding microphone locations.

\subsection{Audio Signal Processing}
At each POE pipeline's initialization, we record a 1.0 second background audio. We average the audio clip's spectrogram with Tukey window of shape parameter 0.25 over time to get the initial acoustic feature vector. In subsequent instances, we collected 0.25 seconds of acoustic signals from each of the six microphones. We then extract spectrograms from each of the acoustic signal clips and average over time. Then we subtract the initial spectrogram feature vector from each microphone's averaged spectrograms and the acoustic features from each of the microphones are concatenated. The concatenated features are then used as the input feature into the rest of the reconstruction pipeline. 

\begin{figure}
    \centering
    \includegraphics[width=1.\linewidth]{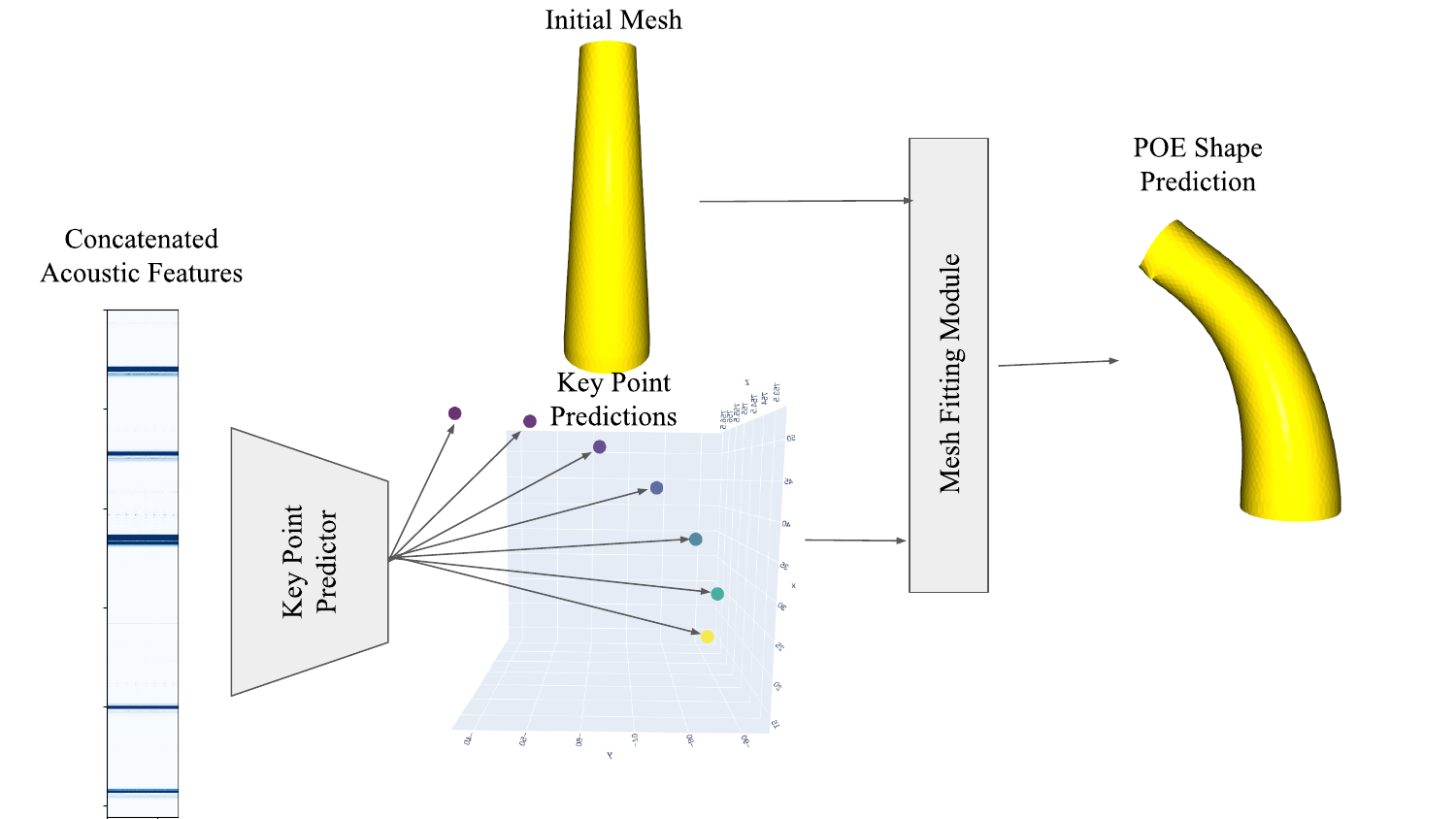}
    \caption{Proposed POE-M Pipeline. First, concatenated acoustic feature vector from six microphones embedded in POE are used to predict new positions of the key points on POE's surface. POE-M uses the known correspondences between the key points and the vertices of POE surface mesh to iteratively fit the mesh to the predicted key points in a physically admissible manner.}
    \label{fig:poem}

\end{figure}
\subsection{POE-M}

We propose POE-Multilayer perceptron (POE-M) reconstruction method. The idea behind POE-M is that by first estimating explicit key point movements from the sensor signals, we can provide strong supervision when we train the acoustic signal decoder and constrain soft robot deformation behaviors to physically admissible transformations. 
\subsubsection{Key Point Model}
Key points provide physically grounded reduced state representation for high degree-of-freedom systems such as fabric \cite{he_fabricfolding_2023} and ropes \cite{lin_softgym_2021}. For POE, using key points is also advantageous because they can be used to provide the proposed POE-M pipeline with stronger supervision during training in contrast to previous works \cite{yoo_toward_2023} that relied on Chamfer distance loss which frequently leads to falling into local optima \cite{li_rearrangement_2023}. 

Another advantage to approximating key points as an explicit intermediate representation of POE state is that POE-M can exploit the key points' physically grounded correspondence to the vertices on POE's surface mesh. This allows POE-M to utilize the predicted sparse key point movements in reconstructing POE's deformed mesh.

In our implementation of the POE-M key point predictor model, the input acoustic feature vector is mapped to the seven 3D point key-point movements with a two-layer MLP. The input acoustic feature from the microphone array has the dimension $512$ and the output is vectorized 3D position changes of the key points.

\begin{figure}
    \centering
    \includegraphics[width=1.6\linewidth]{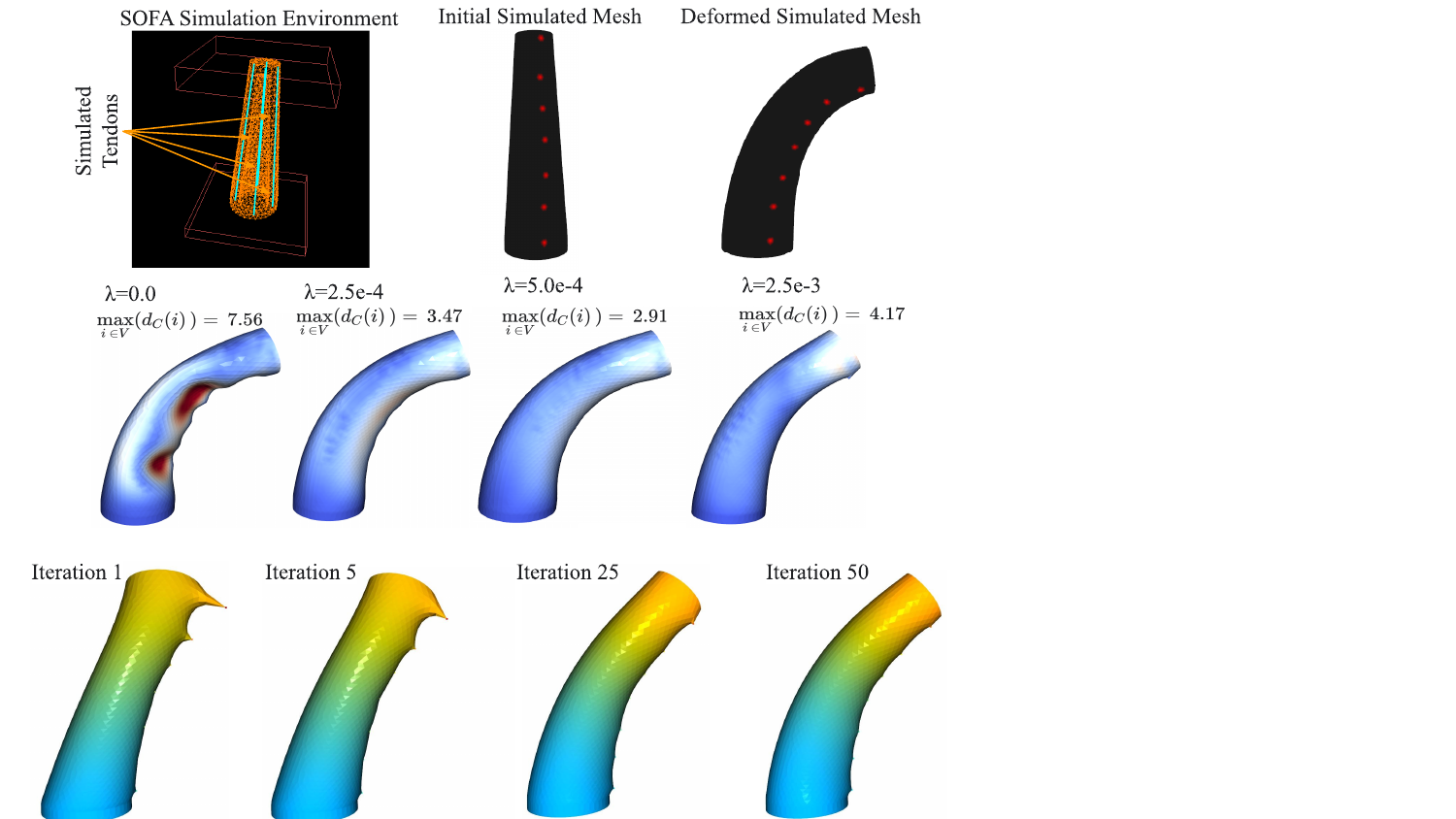}
    \caption{Evaluation of the As-Rigid-As-Possible (ARAP). \textbf{Top row:} SOFA FEM simulation environment \cite{schegg_sofagym_2023} and the generated meshes to be used for ARAP evaluation. \textbf{Middle row:} sensitivity study with varying $\lambda$ parameter. When $\lambda=0$ which corresponds to no neighboring edge rotation regularization, we observe undesirable surface artifacts. All nonzero $\lambda$ parameters removed the artifact effectively where $\lambda=5.0e-4$ yielded the lowest mesh reconstruction error. \textbf{Bottom row:} mesh updates over iterations. After each iteration, the mesh vertices that are not constrained are optimized to reduce the overall ARAP energy. We note that at around 30-50 iterations, the mesh converges. }
    \label{fig:arap}
    \vspace{-5mm}
\end{figure}
\subsubsection{Mesh Reconstruction}
As-Rigid-As-Possible (ARAP) is a method of mesh manipulation popular in animation and graphics applications to generate physically realistic and grounded deformed meshes from a source mesh with chosen handle points that the animators can move to desired positions~\cite{sorkine_as-rigid-as-possible_2007}. We treat the key points' corresponding vertices on the initial POE mesh as the handle points and use the predicted key point positions to infer handle point movements. We first define the source surface mesh $S$ and the deformed mesh $S'$, where each mesh is defined by sets of vertices $V, V'$ and of edges $E, E'$.  We can define the ARAP energy $E_{\mathrm{ARAP}}$ as the following \cite{sorkine_as-rigid-as-possible_2007}:
\[
E_{\mathrm{ARAP}}(S, S') = \sum_{k=1}^{|E|} \min_{R\in SO(3)}\sum_{e_{i,j}\in E} w_{i,j} \| e_{i,j}'- R e_{i,j} \|.
\]

We can then find the solution mesh that minimizes $E_{\mathrm{ARAP}}$ with an iterative local-global optimizer as outlined in Levi, et al. 
 \cite{levi_smooth_2015}. It is a known issue that minimizing $E_{\mathrm{ARAP}}$ with sparse handle points on surface meshes that are moved significantly can result in undesirable surface artifacts. 

 A possible solution to this problem is by minimizing the $E_{\mathrm{ARAP}}$ over a tetrahedral mesh instead of a surface triangular mesh of POE, which implicitly applies soft volumetric constraints that prevent such artifacts from forming. However, because tetrahedral meshes by construction include numerous internal vertices and edges that must be operated over to minimize $E_{\mathrm{ARAP}}$, the process becomes more computationally expensive which is especially undesirable in the context of doing real-time soft robotic proprioception.

 Instead, previous works have demonstrated that the modification of ARAP to include a penalty on the rotations of the neighboring edges produces mesh manipulation that seems physically admissible \cite{levi_smooth_2015}. The new energy to minimize is formulated as
 \vspace{-5pt}
\begin{align*}
     E_{\mathrm{smoothed}}(S, S') = \min_{R_1,..., R_m}\sum_{k=1}^m (\sum_{i,j \in e_k}c_{ijk} \| e_{ij} - R_k e_{ij} \|^2 \\
     + \lambda \hat A \sum_{e_l \in N(e_k)} w_{k l} \| R_k - R_l\|^2).
\end{align*}
We note that minimization of $E_{\mathrm{smoothed}}$ with $\lambda=0$ results in the minimization of $E_{ARAP}$. In POE-M pipeline, vertices corresponding to the key points $p_{1,..., |p_k|}$ are constrained to the new positions based on the predicted key-point positions, and the rest of the mesh vertex positions are moved to minimize $E_{\mathrm{smoothed}}$. As outlined in Fig. \ref{fig:poem}, the key point predictor and the mesh fitting module which utilizes the smoothed ARAP formulation are connected to enable full POE shape prediction from acoustic signal feature vector.
\begin{figure}
    \vspace{-25pt}
    \centering
    \includegraphics[width=1.1\linewidth]{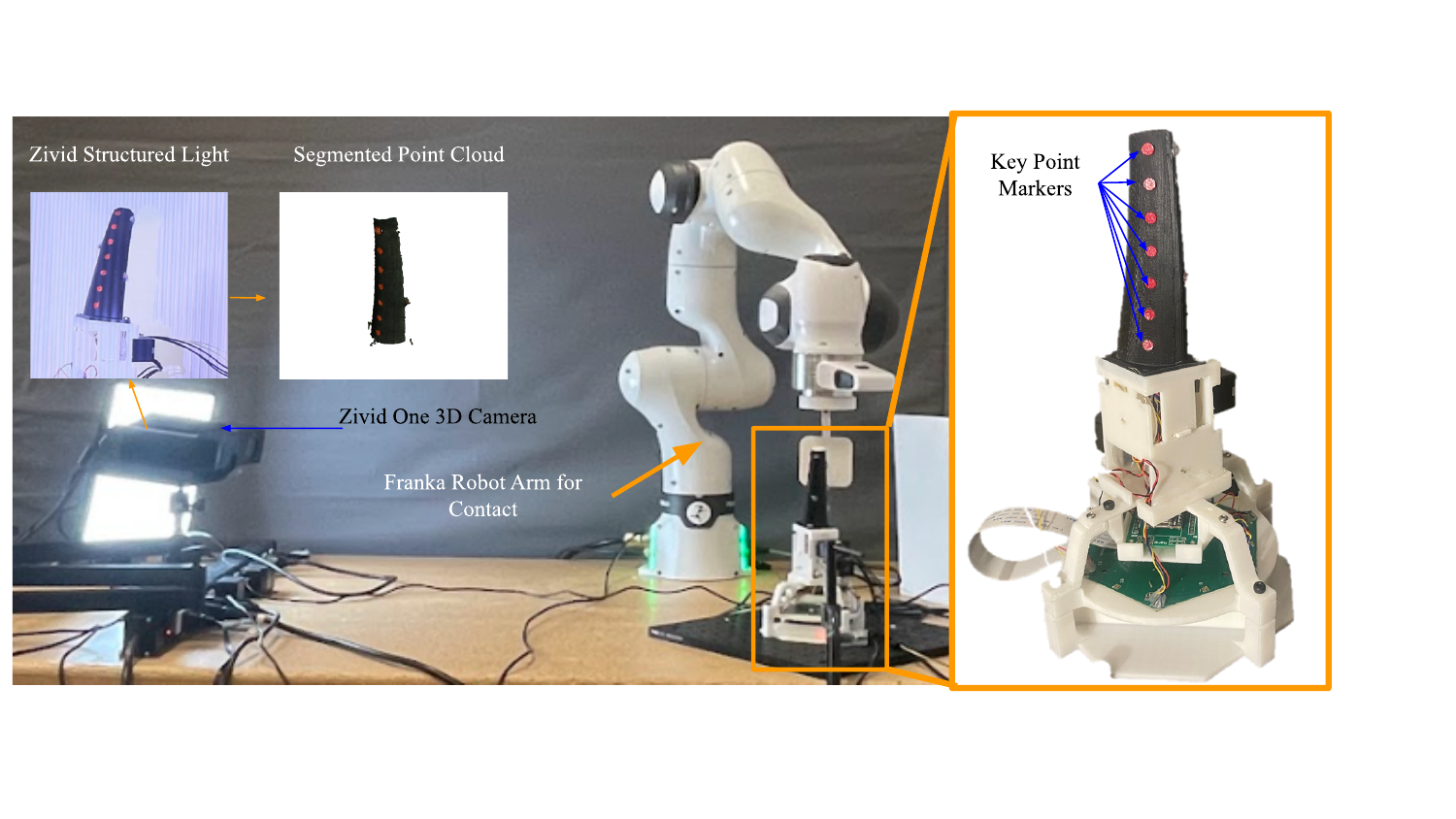}
    \vspace{-25pt}
    \caption{Experimental setup for collecting training and evaluation data. We use a structured light 3D camera to capture high-fidelity point clouds of POE as it deforms. \textbf{Right:} POE has embedded key point markers that are then used to train POE-M to extract key point displacements from acoustic signals. }
    \label{fig:experiment}
\end{figure}

\subsection{Baselines}
As a baseline for acoustic signal, we included a nearest neighbor model that takes in only the two servo motors' encoder positions and match them to the closest seen example and its corresponding observed POE shape. The rest of the POE family of methods utilize acoustic signals from the microphone arrays, each demonstrating that acoustic signal is a viable occlusion-free modality for soft robotic proprioception. First, we present POE-KNN which matches POE acoustic signal to its nearest neighbor in previously seen examples and use the corresponding point-cloud observation as its prediction. We then present POE-DeepSoRo in which we adapt a DeepSoRo decoder-encoder network architecture from vision to acoustics \cite{wang_real-time_2019, yoo_toward_2023}.

\subsection{Experimental Setup}
To evaluate the 3D reconstruction results of the presented POE proprioceptive pipelines and to collect diverse training and evaluation datasets in the real world, we set up the experimental environment as shown in Fig.~\ref{fig:experiment}. The structured light 3D camera (Zivid, One Plus) provides us with low-noise point clouds with 25 $\mu m$ accuracy which serve as a sort of ground truth to evaluate our methods. The 7 degrees-of-freedom robot arm is also installed such that it can create various contact conditions for POE to deform from in addition to POE's tendon-driven range of motion. Such loading conditions allow us to evaluate the various reconstruction pipelines within the expanded configuration space. We collected 5200 point-cloud audio pairs with this setup, and then split 80:20 into training and testing sets respectively. Each point cloud is segmented with HSV color thresholding. For Position-KNN, we also collected servo encoder positions for the two servo motors. For POE-M models, we also segmented the red markers indicating the key point positions on POE. Diffused lighting was added for consistent background illumination and segmentation results. Note that the 3D camera and controlled environment for POE are only required for training and evaluation data collection, and are not requirements for the operation. 
\renewcommand{\baselinestretch}{0.979} 

\section{Evaluation}
\begin{table}
\begin{threeparttable}

\caption{Proprioceptive Performance Evaluation}
\setlength\tabcolsep{0pt} 
\begin{tabular*}{\columnwidth}{@{\extracolsep{\fill}} ll ccc}

\toprule
    Model & Model Input & 
     \multicolumn{2}{c}{Performance Metrics [mm]} \\ 

     & &  Avg. CD $\downarrow$& Max CD  $\downarrow$\\
\midrule
     Position-KNN & Servo Positions  & 5.67 & 20.21 \\ \cdashline{1-4}
          POE-KNN & Audio [All] &   5.93 & 17.02 \\
\addlinespace
    POE-DeepSoRo & Audio [All]  & \bf 4.89 &15.33\\
\addlinespace
     POE-M & Audio [-mic1] & 5.46 & 13.11\\
\addlinespace
     POE-M & Audio [-mic6] & 5.05 & \bf 11.79\\
\addlinespace
     \bf{POE-M} & Audio [All] & 4.91 & \bf 11.98\\
     
\addlinespace

\bottomrule
\end{tabular*}

\scriptsize    
\label{tab:results} 
\end{threeparttable}

\end{table}
\subsection{Simulation Study}
Despite ARAP's wide usage in animation and graphics, its application in modeling real-world mechanics accurately is rarely explored. Furthermore, to the best knowledge of the authors, it has never been applied to modeling deformations of actuated soft robots. The later sections present real-world performance of POE-M and baselines. However, obtaining occlusion-free point cloud of any deforming objects is notably difficult, and disambiguating sensor noise from the method-inherent errors is difficult. To verify that the proposed POE-M's mesh fitting module performs satisfactorily, we test that the module produces results that are aligned with previously tested and verified Finite Element Method (FEM)-based physics simulators. Specifically, we simulated POE mechanics with SOFA framework~\cite{schegg_sofagym_2023} which has been repeatedly demonstrated in literature to match real-world elastic and soft mechanics~\cite{yoo_toward_2023}. 

With the FEM simulator, we generated a deformed mesh of POE under actuation. In the process, we also track how the key points corresponding to the key points on real POE moved. Based on the displacements of the key points, we apply our mesh fitting module on the initial mesh of POE to match it to the deformed mesh. We also study the sensitivity of the $\lambda$ smoothening parameter from $E_{smoothed}$. When $\lambda=0$, $E_{smoothed}$
reduces to minimization of original ARAP energy ($E_{ARAP}$), and we encounter undesirable surface artifacts with a maximum Chamfer distance of 7.56\,mm. With the introduction of a nonzero $\lambda$, there is significant improvement in the Chamfer distance with the best-tested value $\lambda=5\times10^{-4}$, yielding a maximum Chamfer distance of 2.91\,mm.We can observe generally low sensitivity of the mesh-fitting module performance with respect to $\lambda$ values, where each nonzero $\lambda$ value tested performed well. For subsequent experiments, we set $\lambda=5\times 10^{-4}$.

\begin{figure}
    \centering
    \includegraphics[width=1\linewidth]{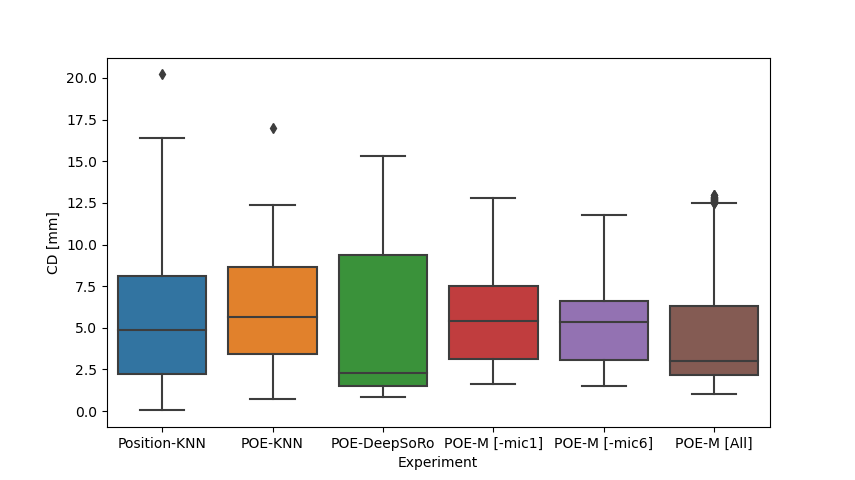}
    \caption{The distribution of unidirectional Chamfer distance across the 1020 evaluation data points. The plot displays outliers marked by diamond markers and the quartiles from the evaluation dataset.}
    \vspace{-15pt}
    \label{fig:box}
\end{figure}
  \begin{figure*}[t!]
  \centering
     \includegraphics[width=1\linewidth]{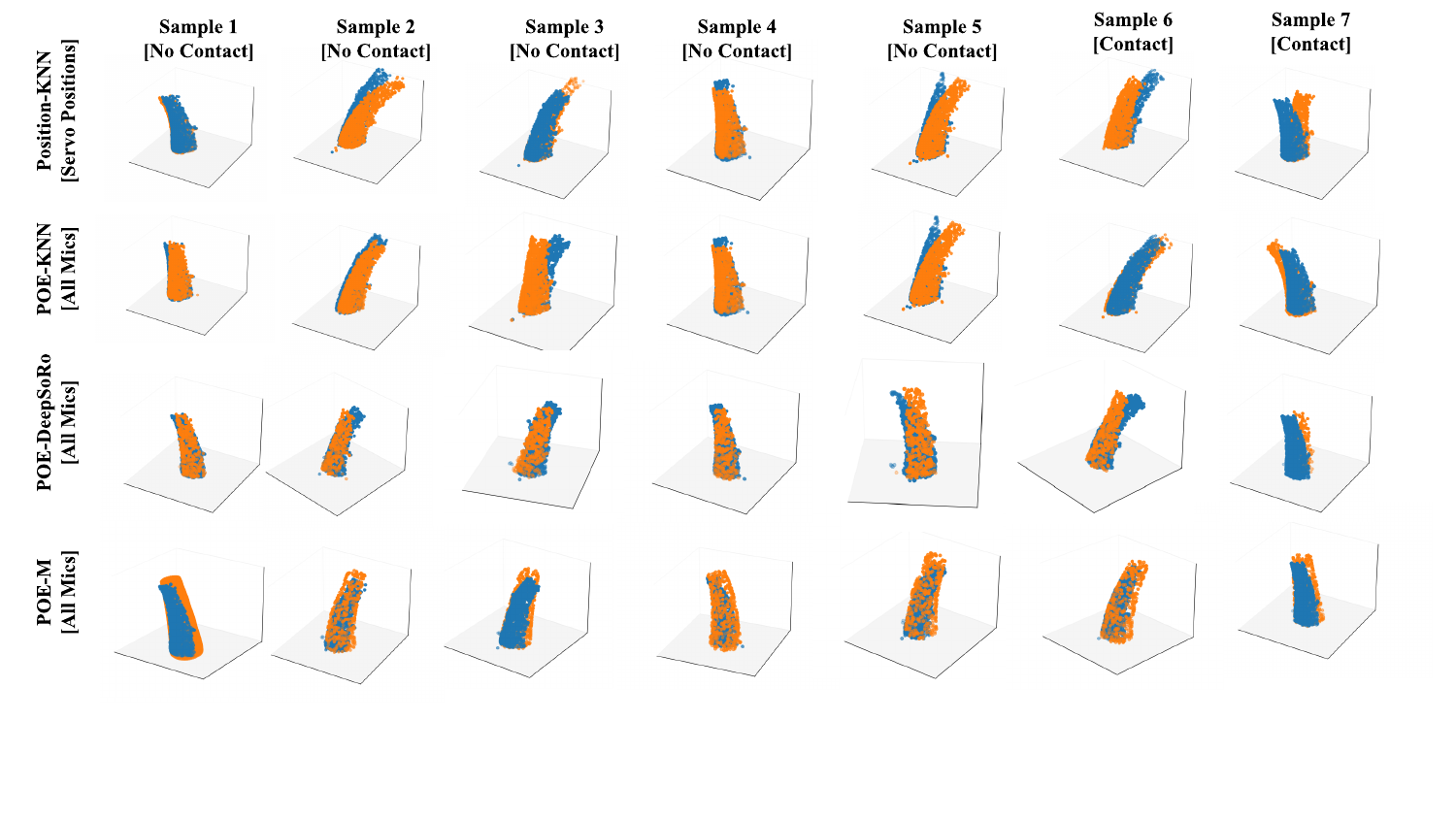}
     \vspace{-53pt}
      \caption{Side-by-side comparison of predicted (orange) and ground-truth observed (blue) point clouds. \textbf{Bottom row}: POE-M uniquely produces a complete point cloud of the deformed POE shape. POE-M also produces generally more stable shape estimates in contrast to POE-DeepSoRo which produces shape estimates with arbitrarily deformed morphology. Both POE-KNN and Position-KNN predictions occasionally fail with large deviations from the ground truth, notable when POE is in contact (samples 6 and 7).}
      \label{fig:results}
      \vspace{-16pt}
\end{figure*}
\subsection{Metrics}
In this work, we propose and report two metrics that must be considered together: average unidirectional Chamfer distance and maximum unidirectional Chamfer distance. The proposed POE-M pipeline generates watertight meshes with dense vertices evenly distributed on POE's finger. However, our real-world experimental setup can only produce a partial point cloud of POE's surface from a single perspective. In practice, this means that many points in the point cloud generated by POE-M will not have corresponding points in the observed ground truth partial point cloud. 

We can define the unidirectional Chamfer distance as 
\begin{equation}
d_{\mathrm{UCD}} (S, T) = \frac{1}{|S|}\sum_{p_i\in S}\min_{p_j \in T} \| p_i- p_j \|_2.
\end{equation}
The intuition is that $d_{\mathrm{UCD}}$ measures how well the observed partial point cloud $S$ fits the target complete predicted point cloud $T$. We report both the $d_{\mathrm{UCD}}$ averaged across different sampled shapes, and the maximum in order to capture both the overall and worst performance.


\subsection{Results}
Table~\ref{tab:results} and Fig.~\ref{fig:box} outlines the performances of both baseline and proposed shape reconstruction methods for POE. Position-KNN baseline method has a comparatively low mean (Table~\ref{tab:results}) and median (centerline of Fig.~\ref{fig:box}). However, the quartile distribution and the maximum recorded Chamfer distance indicate that the baseline method is also prone to fail with commonly large errors ($>$15.0\,mm) and a maximum error of 20.21\,mm. Position-KNN result highlights the primary challenge in proprioception for soft robots, that is, the compliance and underactuation of the soft robots mean that the tendons can not fully constrain the pose. Additionally, tendon-driven robots suffer from various effects of hysteresis and pose-dependent internal tendon friction which alters the soft robot shape even with the same tendon positions ~\cite{y_liu_effect_2021}. We can observe these failures with POE-KNN and Position-KNN in Fig.~\ref{fig:results} where we see large prediction deviations, especially with contact.

As a direct comparison against Position-KNN--which only uses servo encoder positions for soft robot state estimation, the performance of POE-KNN illustrates similarly low average Chamfer distance ($<$ 6.0\,mm) but its range of errors was much tighter with significantly lower maximum Chamfer distance at 15.8\,\% below that of Position-KNN and a sharply lower outlier-removed maximum Chamfer distance as seen in Fig. \ref{fig:box}. Qualitatively, POE-KNN performs well in densely sampled regions as can be seen in Fig.~\ref{fig:results}.

POE-DeepSoRo demonstrates an exceptionally good performance with a low average Chamfer distance ($<$ 5.0\,mm). However, its errors significantly increased in some bending regions, where its Q3 quartile boundary is higher than the baseline methods. This result highlights one of the disadvantages of end-to-end learning for soft robot proprioception in that it learns point-to-point relationships entirely from training examples with no mechanics-based grounding. The result shows that this potentially allows the model to make physically impossible transformations on the point cloud.  

POE-M outperforms the baselines as it preserves the mesh connectivity and returns the full shape estimation of the robot (see Fig.~\ref{fig:results}). This is a crucial advantage of POE-M in that the method can predict even the unobserved surface of POE finger. Furthermore, by framing the soft-robot mesh fitting process as an optimization problem over the edges of the surface mesh, we are instilling stability to the soft robot proprioception problem~\cite{guo_as-rigid-as-possible_2008}. We can note that the results look remarkably consistent with POE's mechanics. These built-in advantages in POE-M are visible in Table~\ref{tab:results} and Fig.~\ref{fig:box} where POE-M has consistently lower variance in error compared to all other methods.

We also present a short ablation study with the microphone signals. When POE-M is retrained with a single microphone signal dropped out, the error remains consistently low. Interestingly, we note from Table \ref{tab:results} that dropping microphone 1's channel seemed to have a much more significant negative impact than dropping microphone 6's channel did. This is potentially explained by the fact that microphone 6 is much further away from the speaker, inducing much weaker signals and perceivable changes as the robot deforms.

\renewcommand{\baselinestretch}{0.979} 

\section{Conclusion}
We present POE, the first soft robotic system with six embedded microphones that enable acoustic-based high-fidelity proprioception. We also present and evaluate a family of methods that utilize these acoustic signals for shape reconstruction and proprioception. Of these, we discuss the benefits of representing states with physically grounded key points and reconstructing the mesh around the key point movements in a physically admissible manner. We quantitatively and qualitatively verified that introducing these soft mechanics-based constraints enables more stable and accurate proprioceptive shape reconstruction results. In future work, we plan to integrate POE-M pipeline into multi-fingered end-effectors for real-time pose estimation and feedback control. In the process, we plan to address the current pipeline's limitation of cross-talk among multiple active sound sources by applying a range of sound source disambiguation methods in literature ~\cite{stefanakis_perpendicular_2017}. Additionally, we plan to improve the system-level integration of POE microphone arrays toward making these pipelines viable for high frequency shape and state estimation of soft robots.



\addtolength{\textheight}{-0.2cm}   


\bibliographystyle{ieeetr}

\bibliography{references.bib}

\newpage

\end{document}